\documentclass[lettersize,journal]{IEEEtran}
\usepackage{amsmath,amsfonts}
\usepackage{array}
\usepackage[caption=false,font=normalsize,labelfont=sf,textfont=sf]{subfig}
\usepackage{textcomp}
\usepackage{stfloats}
\usepackage{url}
\usepackage{verbatim}
\usepackage{graphicx}
\usepackage{cite}
\usepackage{times}
\usepackage{epsfig}
\usepackage{graphicx}
\usepackage{amssymb}
\usepackage{multirow}
\usepackage{listings}
\usepackage{amsthm}
\usepackage{color}
\usepackage{csquotes}
\usepackage{soul}
\usepackage{enumerate}
\usepackage[normalem]{ulem}
\usepackage{algorithm}
\usepackage{algorithmic}
\usepackage{booktabs}
\usepackage{array}
\usepackage{rotating}
\usepackage{arydshln}
\usepackage{ragged2e}
\usepackage{wrapfig}
\usepackage{hyperref}
\hypersetup{breaklinks=true,colorlinks}
\begin{document}

\title{SimLabel: Consistency-Guided OOD Detection with Pretrained Vision-Language Models}
\author{Shu~Zou,
        Xinyu~Tian,
        Qinyu~Zhao,
        Zhaoyuan~Yang,
        Jing~Zhang
\IEEEcompsocitemizethanks{\IEEEcompsocthanksitem Shu~Zou, Xinyu~Tian, Qinyu~Zhao and Jing~Zhang are with School of Computing, the Australian National University, Canberra, Australia. (Email: shu.zou, xinyu.tian, qinyu.zhao, jing.zhang@anu.edu.au).
\IEEEcompsocthanksitem Zhaoyuan~Yang is with GE Research, America. (Email: zhaoyuan.yang@ge.com)
\IEEEcompsocthanksitem Corresponding author: Jing~Zhang.
}
}

\markboth{Journal of \LaTeX\ Class Files,~Vol.~14, No.~8, August~2021}%
{Shell \MakeLowercase{\textit{et al.}}: A Sample Article Using IEEEtran.cls for IEEE Journals}

\maketitle
\begin{abstract}
    Detecting out-of-distribution (OOD) data is crucial in real-world machine learning applications,
    particularly in safety-critical domains. Existing methods often leverage language information from vision-language models (VLMs) to enhance OOD detection by improving confidence estimation through rich class-wise text information. However, when building OOD detection score upon on in-distribution (ID) text-image affinity, existing works either focus on each ID class or whole ID label sets, overlooking inherent ID classes' connection. We find that the semantic information across different ID classes is beneficial for effective OOD detection.
    We thus investigate the ability of image-text comprehension among different semantic-related ID labels in VLMs and propose a novel post-hoc strategy called SimLabel. SimLabel enhances the separability between ID and OOD samples by establishing a more robust image-class similarity metric that considers consistency over a set of similar class labels.
    Extensive experiments demonstrate the superior performance of SimLabel on various zero-shot OOD detection benchmarks.
    The proposed model is also extended to various VLM-backbones, demonstrating its good generalization ability. Our demonstration and implementation codes are available at~\href{https://github.com/ShuZou-1/SimLabel}{SimLabel}.
\end{abstract}

\begin{IEEEkeywords}
OOD Detection, Vision-Language Models.
\end{IEEEkeywords}

\section{Introduction} 
\label{sec:intro}

\IEEEPARstart{H}andling out-of-distribution (OOD) data is
critical
in real-world machine learning applications, particularly in safety-related domains such as autonomous driving systems, open-world recognition and medical diagnosis~\cite{hendrycks2016baseline,Pei2024PSA,Jiang2023Open}. Traditional image domain OOD detection methods primarily focus on visual inputs~\cite{hendrycks2020pretrained, hsu2020generalized, jin2022towards, shen2021enhancing, xu2021unsupervised,sun2024CHF},
where various scoring functions~\cite{wang2022vim, hendrycks2016baseline} are designed to distinguish OOD data from in-distribution (ID) classes. Due to the unimodal nature of these approaches, they rely solely on visual information, limiting their ability to leverage rich semantic information inherent in text labels.

The emergence of Vision-Language Models (VLMs), notably CLIP~\cite{radford2021learning}, has opened new opportunities to leverage paired image and text information for OOD detection. For instance, ZOC~\cite{Esmaeilpour_2022} utilizes a trainable captioner to generate OOD labels and introduces the task of Zero-Shot OOD detection, which does not require training on ID samples. Maximum Concept Matching (MCM)~\cite{ming2022delving} proposes a distance-based zero-shot OOD detection method where the fundamental assumption is that images are more likely to be ID if their embeddings are closer to ID text embeddings, and vice versa. However, the naive score construction in this method neglects the rich semantic textual information of the ID classes, leading to less effective ID/OOD separation.

\begin{figure*}
    \centering
    \includegraphics[width = 0.9\textwidth]{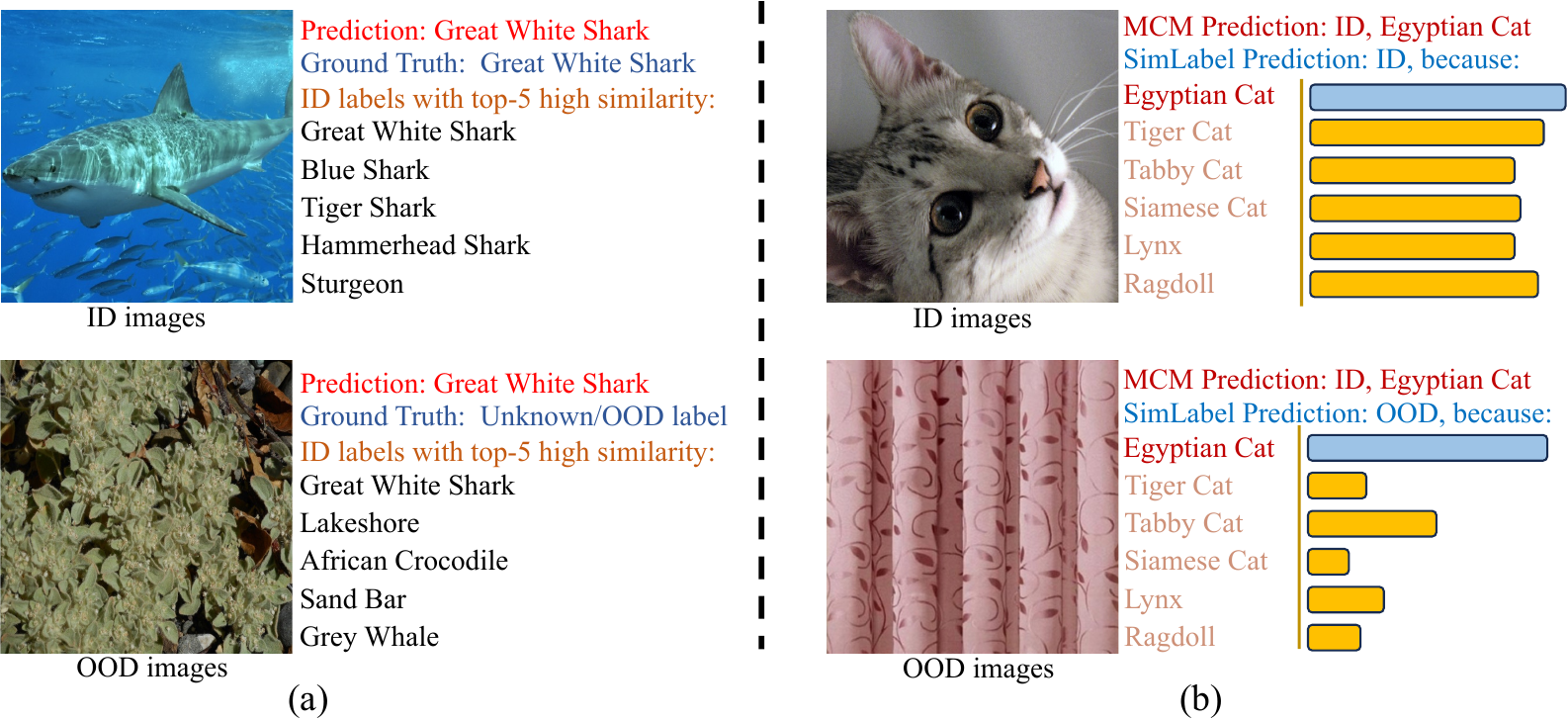}
    \caption{(a) Illustration of VLMs guided OOD detection for
    ID (top image from ImageNet~\cite{imagenet2009}) and OOD (bottom image from iNaturalist~\cite{vanhorn2018inaturalist}) samples, respectively.
    (b) Comparison between the proposed SimLabel and the baseline MCM~\cite{ming2022delving} for ID (top) and OOD (bottom) samples, demonstrating how our method detects by aggregating OOD scores across similar ID labels (yellow and blue bars denote image \& similar-classes-labels similarity and image \& class-labels similarity respectively).
    }
    \label{fig:motivation_sim_class}
\end{figure*}

To address these limitations, variants of the MCM score have been presented. For instance, ~\cite{dai2023exploring} introduces class-wise attributes to enhance the confidence score between ID images and labels, providing more accurate and expressive descriptions for improved performance. Similarly, ~\cite{wang2023clipn}
trains a negative prompt for each ID class using an external dataset, performing OOD detection by combining scores from both negative and traditional prompts. Recent research~\cite{jiang2024neglabel} presents
negative text labels which assumes the ID images show high-similarity to the whole set of ID labels. However, these methods either primarily focus on learning individual class-wise textual information or depends on introducing external negative prompts, overlooking the semantic information existing among different classes.
In Fig.~\ref{fig:motivation_sim_class} (a), we show OOD detection results for ID (top) and OOD (bottom) samples, respectively, without considering intra-class similarity. 
Our basic assumption is that an ID sample should consistently have high similarity scores across similar ID classes. 
This assumption motivates us to devise a new method for detecting OOD samples based on measuring the consistency among semantically related labels from the ID classes.


To further illustrate our motivation, we conduct a straightforward experiment verifying assumption in Sec~\ref{sec:motivation}. Based on the observation, for every ID label, we aim to introduce a set of accompanying or similar labels from either the ID labels or external knowledge that enable the model to detect OOD samples in a second direction: consistency between image and similar classes. Specifically, as demonstrated in Fig.~\ref{fig:motivation_sim_class} (b), given ID (top) and OOD (bottom) images which are predicted as the same class by the baseline method, ID images show consistent higher similarity to the set of semantically similar ID classes than OOD images.
Based on this observation, we propose SimLabel, a post-hoc method with a well-designed OOD score to detect OOD images by examining the consistency of high-similarity over similar classes.
For instance, Fig.~\ref{fig:motivation_sim_class} (b) illustrates the detection on the OOD sample as it receives diverse similarity scores for similar classes, namely Tiger Cat, Tabby Cat, Siamese Cat etc.
The proposed OOD score combines knowledge from the prediction and its similar classes, thus better leveraging the VLMs’ capabilities of comprehending class prototypes. 

Additionally, we design several algorithms for selecting high-quality similar labels from the ID class or external knowledge. The choice of similar classes can be generated from three directions: text-hierarchy, world knowledge, and pseudo-image-label alignment. Extensive experiments validate that our proposed method, SimLabel, achieves superior performance across various zero-shot OOD detection benchmarks. 

We summarize our main contributions as follows:


\begin{itemize}
\item We propose a novel post-hoc framework, SimLabel (see Sec.~\ref{Sec:SimLabel}), that constructs the affinity between images and class prototypes with semantic-related labels for robust OOD detection.
\item We introduce different and comprehensive strategies (see Sec.~\ref{sec:choise_Sim_Class}) for selecting similar labels from the various perspectives and illustrate the influence on OOD detection performance with the different choices of similar classes (see Sec.~\ref{Sec:experiment}).
\item  We present in-depth empirical analysis, offering insights into the effectiveness of the SimLabel score (see Sec.~\ref{Sec:analysis}) and show that SimLabel learns a robust and discriminative image-class matching score, potentially improving visual classification ability.

\end{itemize}
\section{Related works}
\textbf{Out-of-Distribution Detection.} 
Conventionally, the objective of OOD detection is to derive a binary ID-OOD classifier to detect OOD images within the test dataset. Specifically, without modifying the network architecture, the key factor is finding the different patterns between the output of ID and OOD samples, making the designing the OOD score as one of the most important tasks in OOD detection. We can mainly separate the existing methods in to three types of OOD scores, namely the probability-based, logit-based, and feature-based, respectively. MSP~\cite{hendrycks2016baseline} uses the maximum predicted probability as the score and~\cite{liang2020enhancing} aims to get rid of the over-confidence through perturbing the inputs and re-scaling the logits. MaxLogits~\cite{hendrycks2022scaling} utilizes the maximum of logits as the score and Energy~\cite{liu2021energybased} defines
the energy-function as the OOD score. ReAct~\cite{sun2021react} and DICE~\cite{sun2022dice} further investigate the improvement of energy score through the feature clipping and discarding. Among the feature-based methods, Lee~\cite{lee2018simple} propose the score via the measurement of minimum Mahalanobis distance between the feature and the class-wise centroids as the OOD score. KNN~\cite{sun2022knn} investigates the effectiveness of non-parametric nearest-neighbor distance for OOD detection. 

\textbf{OOD Detection with Vision-Language Representations.}
With the rise of large-scale pre-trained VLMs, there are various works focusing on utilizing textual information for visual OOD detection. For instance, \cite{fort2021exploring}
proposes the utilization of VLM for OOD detection through the generation of the candidate OOD labels. MCM~\cite{ming2022delving} is a conventional post-hoc zero-shot method that uses the maximum predicted softmax value as the OOD score for OOD detection. Based on MCM, NPOS~\cite{tao2023nonparametric} conducts the OOD data synthesis and fine-tunes the image encoder to find a decision boundary.
Dai et al.~\cite{dai2023exploring} introduces class-wise attributes to enhance the confidence score between ID images and labels while overlooking the semantic information among different ID labels. CLIPEN~\cite{dai2023exploring} investigates the positive and negation-semantic prompts in separating ID and OOD domain.
NegLabel~\cite{jiang2024neglabel} introduces external negative labels to enhance ID/OOD separation. 

Similar to both CLIPEN~\cite{dai2023exploring} and NegLabel~\cite{jiang2024neglabel}, we also propose method using external text knowledge to boost OOD detection tasks. Differently, our main method (SimLabel-I) aims in building a more reasonable and robustness ID image-text pairing with extensive similar ID classes exploration. Particularly, we detect samples through measuring consistency between images and similar classes, enabling the model to build affinity from images to various ID labels in an interpretable and robust manner which enhances ID/OOD separation.

\section{Preliminaries}\label{Sec:preliminaries}

Let $ \mathcal{X}=\mathcal{X}_{ID}\cup \mathcal{X}_{OOD}$, $\mathcal{L}=\left\{ l_1,\dots l_C \right\}$ and $\mathcal{P}=\{{\rm prompt}(l_c)\mid l_c\in\mathcal{L}\}$ be the set of images, ID labels and corresponding prompts respectively
where $c$ indexes ID class.
The function $\textrm{prompt} (l_c)$ denotes the prompt template, e.g., \enquote{A photo of [class]}.
We define ID images as $x_{\textit{ID}} \in \mathcal{X}_{ID}$ and OOD images as $x_{\textit{OOD}} \in \mathcal{X}_{OOD}$.

\textbf{CLIP model~\cite{radford2021learning}.}  Given any images $x \in \mathcal{X}$ and label $l_x \in \mathcal{L}$, along with frozen text encoder $f_T:\mathcal{X} \to \mathcal{R}^D$ and image encoder $f_I:\mathcal{P} \to \mathcal{R}^D$ from the CLIP model, respectively, the visual features $\mathbf{h} \in \mathcal{R}^{D}$ and the textual feature $\mathbf{e}_c \in \mathcal{R}^{D}$ can be extracted as:
\begin{equation}\label{equa:features_extraction}
    \mathbf{h} = f_I(x), \mathbf{e}_c  = f_T( \textrm{prompt} (l_c)),
\end{equation}
where $D$ denotes the dimension of features. The CLIP model performs prediction through the measurement of cosine similarity between image embedding
$\mathbf{h}$ and text embedding
$\mathbf{e}$. Thus, the prediction can be selected as the label $\hat{l}$ with highest similarity $\mathcal{M}(x,\hat{l})$, expressed as:
\begin{equation}\label{equa:CLIP_classify}
\begin{aligned}
    \hat{l} &= \underset{l_c\in \mathcal{L}}{\arg\max} \left\{ \mathcal{M}(x,l_c)\right\} \\
    \mathcal{M}(x,l_{c}) &= \cos(f_{I}(x),f_T(\textrm{prompt} (l_c)).
\end{aligned}
\end{equation}

\textbf{Score function.}  Score function plays an essential role in OOD detection tasks. Given score function $S(\cdot)$ along with the threshold $\tau$, following many representative works in OOD detection~\cite{hendrycks2016baseline,ge2023improving,ming2022delving}, an image $x$ can be decided as ID or OOD based on
function $G$:
\begin{equation}\label{equa:ood_traition}
    G_{\tau}(x)=\left\{\begin{array}{ll}
    \textrm{ID} & S(x) \geq \lambda \\
    \textrm{OOD} & S(x) <\lambda
\end{array}.\right.
\end{equation}

The performance of OOD detection is highly related to the design of function $S(\cdot)$, where ID samples are expected to receive higher scores than the OOD samples.

\textbf{Problem set-up.} VLMs bridge image and text modalities through a pair-matching training strategy. Thus, given a pre-trained CLIP-like~\cite{radford2021learning} model and pre-defined label names, one can conduct visual classification without training (namely zero-shot classification task). In this paper, we follow the setting of this task, aiming to develop a score to detect any input
that does not belong to any ID classes without sacrificing the classification accuracy.

\section{Methodology}\label{Sec:method}

\subsection{CLIP-based OOD detection}
Leveraging its extensive training volume and large model size, CLIP has demonstrated remarkable generalization capabilities in zero-shot classification tasks. This raises the question of its potential effectiveness in zero-shot OOD detection, which merits further exploration. To begin, we review the methodology for performing zero-shot OOD detection based on existing studies~\cite{ming2022delving,jiang2024neglabel}.
For a given set of ID classes, 
Image-text similarity is computed with Eq.~\ref{equa:CLIP_classify}, functioning analogously to a traditional classifier. Building on the widely used MSP method~\cite{hendrycks2016baseline}, MCM~\cite{ming2022delving} adopts the maximum similarity between an input image and text labels as the OOD score, based on which OOD samples are decided with Eq.~\ref{equa:ood_traition}.

\begin{figure}
    \centering
    \includegraphics[width = 0.49\textwidth]{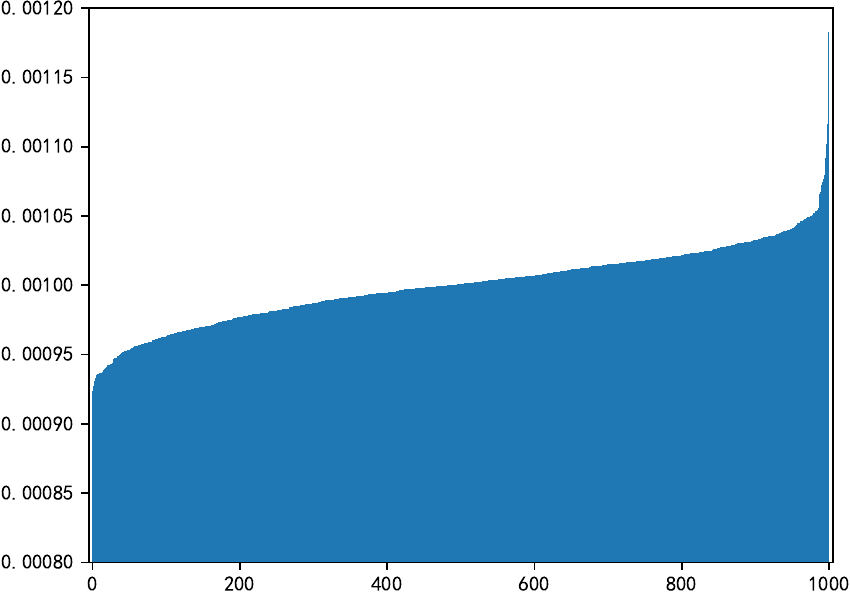}
    \caption{This figure demonstrate the sorted average similarity between a specific class of images
    with the whole label set (1,000 labels in our case).
    Images $x \in \mathcal{X}$ show high similarity to several ID classes rather than one single label. Notably, we follow the score design in MCM~\cite{ming2022delving} where the similarity are transformed with the Softmax function.}
    \label{fig: motivation}
\end{figure}

Existing MSP-based methods~\cite{ming2022delving, dai2023exploring} rely solely on the label with the highest similarity, where this \enquote{label} is referred to as the pseudo-label. This approach restricts the utilization of CLIP's full image-text comprehension capabilities. Prior studies~\cite{ge2023improving, beyer2020we} reveal the
multi-label ambiguity issue and find that
the sensitivity of text encoders leads to the failure cases in CLIP's image-text understanding, potentially limiting its performance in OOD detection.
To address these limitations, self-consistency methods~\cite{wang2023selfconsistency, ge2023improving} have been shown to effectively enhance reasoning accuracy by incorporating multiple prompts. Building upon this idea, we propose extending self-consistency to the domain of confidence estimation for multi-modal OOD detection. Specifically, we measure self-consistency across different class labels to improve the robustness and reliability of OOD detection using CLIP.

\begin{figure*}
    \centering
    \includegraphics[width = 0.8\textwidth]{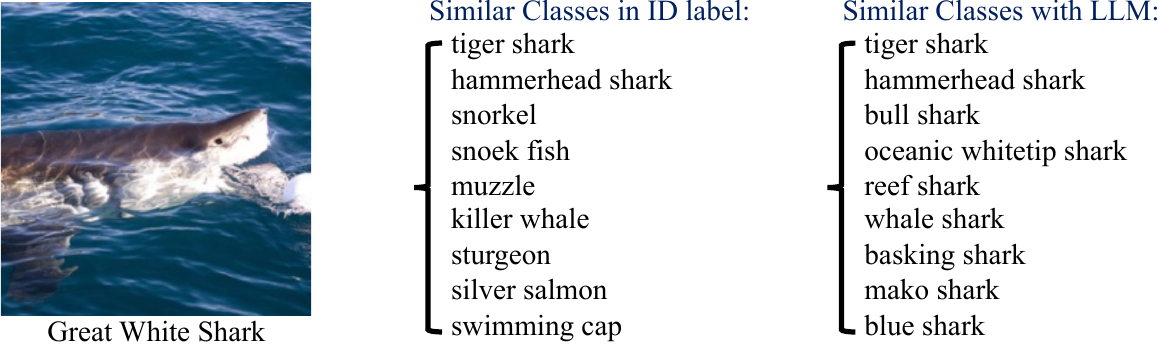}
    \caption{This figure illustrates samples of similar classes for the class "Great White Shark" using methods in Sec.~\ref{sec:Co-Ocu_sim_class_generation} and Sec.~\ref{sec:similar_class_GPT}. \textbf{Left} The similar classes generated from the ID labels. \textbf{Right} The similar classes generated by LLM.}
    \label{fig: Similar Classes Generated Visual}
\end{figure*}

\subsection{Motivation}\label{sec:motivation}

This study is motivated by the identification of a critical yet often overlooked issue: the ID image-text similarity score serves as an useful indicator in explaining
the intrinsic connections among different classes. 
In particular, when assessing the similarity between images and ID labels, especially within large-scale datasets such as ImageNet~\cite{imagenet2009}, the image-text similarity score
offers valuable insights into the text comprehension capabilities of Vision-Language Models (VLMs). We thus design a
straightforward experiment to explore the existence of similar classes.
Specifically, given an image set $\mathcal{X}$ and its corresponding label set $\mathcal{L}$, we compute the similarity between each image and every label using Eq.~\ref{equa:CLIP_classify}. For randomly selected images $x \in \mathcal{X}_i$ where their labels are $l_i$, we calculate the average similarity between the images and each text label. The sorted similarity distribution is visualized in Fig.~\ref{fig: motivation}.
As depicted in Fig.~\ref{fig: motivation}, images in $\mathcal{X}$ generally exhibit high similarity to a small subset of ID labels, including but not limited to the ground truth label. This finding underscores the presence of similar classes, highlighting inherent connections among ID labels. By leveraging the generated class-wise similar classes, we can further determine whether an image is ID
in-distribution (ID) 
or OOD
based on this consistency measurement.

\begin{table}[h!]
\setlength{\tabcolsep}{10mm}
  \begin{center}
    \caption{Subset of ImageNet Class Hierarchy}
    \label{tab:hierarchical-set}
    \begin{tabular}{lc}
      \textbf{Super-Class} & \textbf{ID-Labels/Child-Classes}\\
      \hline
      \multirow{3}{*}{Shark} & Hammerhead Shark \\ 
      & Great White Shark \\ 
      & Tiger Shark \\
      \hline
      \multirow{2}{*}{Flower} & Daisy \\ 
      & Orchid \\ 
      \hline

      \multirow{4}{*}{Turtle} & Mud Turtle \\ 
      & Terrapin \\ 
      & Box Turtle\\
      & Sea Turtle\\
      \hline
      \multirow{5}{*}{Domestic Cat} & Tabby Cat \\ 
      & Tiger Cat \\ 
      & Persian Cat\\
      & Siamese Cat\\
      & Egyptian Cat\\
      \hline
      \multirow{1}{*}{Reservoir} & Water Tower \\
     \bottomrule
    \end{tabular}
  \end{center}
\end{table}
\subsection{Similar class generation}\label{sec:choise_Sim_Class}

\textbf{Overview of similar classes generation.} We aim to generate a set of labels for each class $l_c$ that have higher affinity/similarity to ID samples compared with OOD samples.
We then refer these labels as similar labels $\mathcal{D}(l_c)$ for class $l_c$.
In this section, we propose three methods for generating similar classes: \textbf{1.} explore the text hierarchy among class labels and select the labels under the same super-class as similar classes (see Sec.~\ref{sec:Naive_sim_class_generation}); \textbf{2.} use the external large language models/world knowledge in generating similar classes (see Sec.~\ref{sec:similar_class_GPT}); \textbf{3.} utilize the similarity between ID images and ID labels and pseudo-labeling method for selecting the similar classes (see Sec.~\ref{sec:Co-Ocu_sim_class_generation}).

\subsubsection{Similar classes based on text hierarchy }\label{sec:Naive_sim_class_generation}
Utilizing inherent hierarchical information among different class labels is a natural idea in clustering ID classes into groups with high semantic correlation and ~\cite{novack2023chils} has shown a precise construction of tree-structured hierarchical label sets over various datasets. There has been extensive researches in studying the hierarchical connection~\cite{krizhevsky2012imagenet,bilal2017convolutional,tempfli2024hiernet} for image understanding.  Here, we follow the pipeline shown in~\cite{novack2023chils} to build hierarchical label sets among ID labels where the subsets of the hierarchy can be seen in Table~\ref{tab:hierarchical-set}, we select the class labels under the same super-class as the set of similar classes \(\mathcal{D}(l_c)\). We denote the SimLabel score with these hierarchical similar classes as SimLabel-H.

\subsubsection{Similar classes from large language models}\label{sec:similar_class_GPT}
Nevertheless, the above method may face with the problem of unbalanced ID label space which result in the lack of similar labels for some rare ID classes. Large language models (LLMs), such as GPT-3~\cite{brown2020language}, possess extensive world knowledge across various domains, leading to
a direct approach for generating similar classes with LLMs. We prompt LLMs to generate similar classes $\mathcal{D}(l_c)$ where they share similar visual features. We randomly select several visual categories and manually compose similar classes using a one-shot in-context example. 
Specifically, we employ the following query template:
\noindent\fbox{
\parbox{0.46\textwidth}{
Given a specific class label, generate a list of visually similar class labels. 

Here are some examples to illustrate how you should structure your answers:

Given class label: CD player

Your answer: tape player, cassette player, radio, cassette, modem, desktop computer, monitor, hard disc, remote control, loudspeaker

Given class label: {category}

Your answer:
}
}

An example of generated similar classes is listed in Fig.~\ref{fig: Similar Classes Generated Visual}.
Additionally, although the LLM generates semantically correlated similar classes from world knowledge, we find that some classes are more semantic-related and potentially affect the OOD detection performance. We measure the semantic textual similarity between $l_c$ and every similar label $\mathbf{d} \in \mathcal{D}(l_c)$ with cosine similarity in feature space, and select the labels with top-$k$ similarities as the final set of similar classes. We denote the SimLabel score with prompting LLMs as SimLabel-L.


\subsubsection{Similar classes with image-text alignment}\label{sec:Co-Ocu_sim_class_generation}
The selection of similar classes is to find labels whose ID samples have high affinity.
The above two selection methods are limited to the textual modality, where the sensitivity of the text encoder~\cite{miyai2023zeroshot,he2019hard} may result in the incorrect selection of similar classes.
In this section, we introduce a new strategy for selecting similar classes with consideration of ID image-text alignment. For each ID image, we can first perform zero-shot visual classification to assign it to a pseudo ID class and select the labels with top similarities as similar classes.

The generation of similar classes through single-image-text alignment can be problematic due to CLIP's inaccuracy and contingency in image-text alignment. To address this issue, we propose a robust method to select similar classes that consistently show high similarity among most ID samples $\mathcal{X}_c$ predicted into class $l_c$. Specifically, our assumption is that, for every image $x_c \in \mathcal{X}_c$, set of similar class (donates $\mathcal{D}(x_c)$) with top-$k$ similarity varies but the true similar classes representing class prototype $l_c$ will consistently or highly possibly show in $\mathcal{D}(x_c)$.
In this case, we record all set of similar labels  $\mathcal{D}(x_c)$ for each images in $\mathcal{X}_c$ and select the labels with top occurrence among all sets as the similar classes $\mathcal{D}(l_c)$ for class prototype $l_c$. The detail of algorithm is shown in Algorithm~\ref{alg:generate_simclass}. We denote the SimLabel score by referring to image-text alignment as SimLabel-I.

\begin{figure*}
    \centering
    \includegraphics[width = 1.0\textwidth]{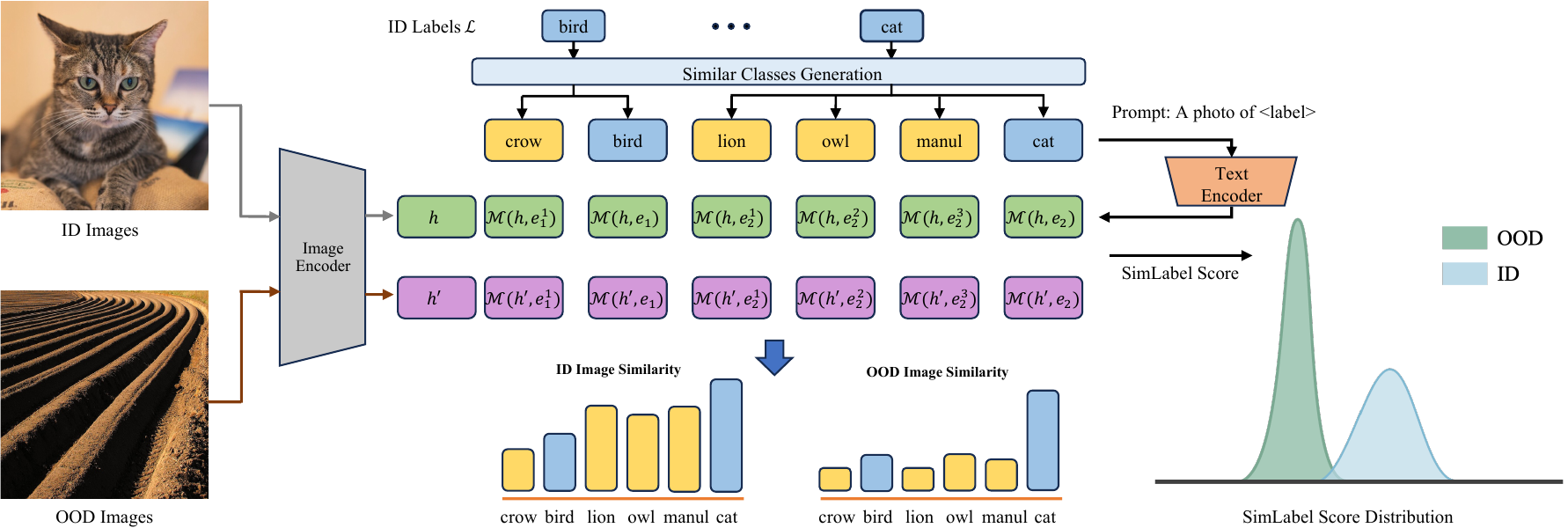}
    \caption{\textbf{Overview of the SimLabel zero-shot OOD detection framework.} The image encoder first encodes ID and OOD images into image embeddings $\mathbf{h}$ and $\mathbf{h}'$, respectively. For every class label (represented as blue blocks) in the ID label set $\mathcal{L}$, similar classes (represented as yellow blocks) are generated through the process of similar class generation defined in Sec.~\ref{sec:choise_Sim_Class}. The text encoder extracts ID and their similar class labels into text embeddings with prompts and the image-text similarities are measured using the function defined in Eq.~\ref{equa:CLIP_classify}. Both image and text encoders are frozen. The below charts indicate that, ID $\textit{cat}$ images, compared with OOD images that are predicted into the $\textit{cat}$,  will produce higher similarity to similar classes such as $\textit{lion, owl, manul}$. Our proposed SimLabel (detailed in Sec.~\ref{Sec:SimLabel}) conducts OOD detection by utilizing image \& class-label and image \& similar-classes-label similarity.} 
    \label{fig:inference_pipeline}
\end{figure*}

\begin{algorithm}[t]
\caption{Similar Class Generation with Class-wise Image-Text Similarity}
\label{alg:generate_simclass}
\begin{algorithmic}[1]
\REQUIRE ID label set $\mathcal{L}$, ID sample $x_{ID} \in \mathcal{X}_{ID}$
\ENSURE Similar Classes $\mathcal{D}(l_c)$ for every $l_c \in \mathcal{L}$

\FOR{$l_c \in \mathcal{L}$}
    
    \STATE $\mathcal{X}_{c} \subset \mathcal{X}_{ID}$
    \STATE $//$ Subset of $\mathcal{X}_{ID}$ with Eq.~\ref{equa:CLIP_classify} whose pseudo label is $l_c$
    \STATE $\mathcal{D}(\mathcal{X}_c) \gets \emptyset$ 
    \FOR{$x_c \in \mathcal{X}_c$ }
        \STATE Compute $\mathcal{D}(x_c)$ 
        \STATE $//$ Find ID labels with top-$k$ high-similarity for $x_c$
        \STATE $\mathcal{D}(\mathcal{X}_c)$ records all the similar classes $d(x_c) \in \mathcal{D}(x_c)$
    \ENDFOR
    \STATE $\mathcal{D}(l_c) \gets  \textit{Select} (\mathcal{D}(\mathcal{X}_c)$) 
    \STATE $//$ Select labels in $\mathcal{D}(x_c)$ with top-$k$ highest occurrence
\ENDFOR
\RETURN{} $\mathcal{D}(l_c)$ for every $l_c \in \mathcal{L}$ 
\end{algorithmic}
\end{algorithm}



\subsection{OOD detection with similar classes}\label{Sec:SimLabel}

\textbf{SimLabel Score.} 
With the generation of high-quality similar-classes, in this section, we propose our pipeline in using image \& similar-class-label similarity for detecting OOD sample as illustrated in Fig.~\ref{fig:inference_pipeline}. For every class, given the class-wise similar classes $\mathcal{D}(l_c)$ generated in Sec.~\ref{sec:choise_Sim_Class}, we merge them with the class label $l_c$ to obtain an extended class-wise label set.
Then, the extended labels set are fed into the text encoder to obtain text embeddings as shown in the yellow bar in Fig.~\ref{fig:inference_pipeline} and CLIP calculates the cosine similarities between the text and image embeddings. Formally, we define the affinity $\mathcal{A}(x,l_c)$ between images $x$ and class $l_c$ as:
\begin{equation}\label{equa:affinity_simclass}
    \mathcal{A}(x,l_c) = \mathcal{M}(x,l_{c}) + \alpha *\underset{d \in \mathcal{D}(l_c)}{\sum} \mathcal{M}(x,d)/|\mathcal{D}(l_c)|
\end{equation}
where $|\mathcal{D}(l_c)|$ indicates the cardinality of similar classes and $\alpha$ is a hyper-parameter that determines the weight of image \& similar-classes-label similarity. The higher the $\alpha$ is, the impact of similar classes in SimLabel score will be amplified. Intuitively, Eq.~\ref{equa:affinity_simclass} enhances the estimation of connections between image and class prototype with the combination of image \& class-label and weighted image \& similar-classes-label similarity.
Motivated by the assumption in~\cite{ming2022delving} that the maximum similarity of ID image-text alignment shows advantages over OOD samples, we formally define our SimLabel score with the maximum matching score as:
\begin{equation}\label{Equa:simlael_equation}
    S(x;\mathcal{L},\tau) =\underset{l_c\in \mathcal{L}}{\max} \frac{ e^{\mathcal{A}(x,l_c)/\tau }}{\sum_c^{C} e^{\mathcal{A}(x,l_c)/\tau }}
\end{equation}
where $\tau$ indicates the temperature scalar. Our OOD detection function can then be formulated as:
\begin{equation}
    G(x;\mathcal{L},\tau)=\left\{\begin{array}{ll}
    \textrm{ID} & S(x;\mathcal{L},\tau) \geq \lambda \\
    \textrm{OOD} & S(x;\mathcal{L},\tau) <\lambda
\end{array},\right.
\end{equation}
where $\lambda$ is chosen so that a high fraction of ID data (e.g., 95\%) is above the threshold. For sample $x$ that is classified as ID, one can obtain its class prediction based on the nearest prototype: $\hat{y} = \underset{l_c \in \mathcal{L}}{\arg \max} \mathcal{A}(x,l_c)$.


\section{Experiment}\label{Sec:experiment}

\subsection{Experiment setup}\label{sec:experiment-setup}
\textbf{Datasets and benchmarks.} We evaluate our method on the ImageNet-1k OOD benchmark~\cite{huang2021importance} and primarily compare it with the MCM method~\cite{ming2022delving} due to its promising and consistent performance in the zero-shot OOD detection task. The ImageNet-1k OOD benchmark
uses the large-scale visual dataset ImageNet-1k as ID data and iNaturalist~\cite{vanhorn2018inaturalist}, SUN~\cite{SUN}, Places~\cite{zhou2016places}, and Texture~\cite{cimpoi2013describing} as OOD data, covering a diverse range of scenes and semantics. Each OOD dataset has no classes that overlap with the ID dataset.

\textbf{Implement details.} In our experiments, we adopt CLIP~\cite{radford2021learning} as the target pre-trained model, which is one of the most popular and publicly available VLMs. Note that our method is not limited in CLIP, and it can be applicable to other vision-language pre-trained models that enable multi-modal feature alignment. Specifically, we provides additional experiments in investigating the effectiveness of SimLabel based on various VLM architectures seeing Sec.~\ref{sec:dis} for details.
Our experiments are primarily conducted using the CLIP-B/16 model, which consists of a ViT-B/16 Transformer as the image encoder and a masked self-attention Transformer~\cite{attention2017} as the text encoder. 

We propose three different strategies for similar classes generation. Particularly,
to select similar classes in SimLabel-H, we follow the construction of hierarchical label sets in~\cite{novack2023chils}, and obtain accurate super-classes for the ImageNet labels.
The LLMs we prompt for SimLabel-L is the leading proprietary model GPT-4~\cite{Achiam2023GPT4TR}. In generating similar classes for SimLabel-L and SimLabel-I score, we select the quantity of similar classes with top-$k$ similarities, where we have $k=6$ in this paper. Additionally, following the theoretical analysis and setting in~\cite{ming2022delving}, we set temperature $\tau = 1$ in our matching score in Eq.~\ref{Equa:simlael_equation}. Empirically, we set the weight of image \& similar-classes-label $\alpha = 1$ in Eq.~\ref{equa:affinity_simclass}. All experiments are conducted on a single NVIDIA 4090 GPU.

\textbf{Metric.} For evaluation, we mainly use two metrics: (1) the false positive rate (FPR@95) of OOD samples when the true positive rate of in-distribution samples is 95\%, and (2) the area under the receiver operating characteristic curve (AUROC).

\subsection{Experimental results and analysis}
\begin{table*}[t]
\resizebox{1\textwidth}{!}{\begin{minipage}{1\textwidth}
\centering
\setlength{\tabcolsep}{3.3mm}
\renewcommand{\arraystretch}{1.1}
\caption{OOD detection performance comparison with baselines on ImageNet-1k benchmark using CLIP-B/16 model. SimLabel-H, SimLabel-L and SimLabel-I indicate the SimLabel score using similar labels generated in Sec.~\ref{sec:Naive_sim_class_generation}, Sec.~\ref{sec:similar_class_GPT},Sec.~\ref{sec:Co-Ocu_sim_class_generation} respectively.}
\label{table:imagenetOOD}
\begin{tabular}{lcccccccccc}
\bottomrule

\multicolumn{1}{c}{\multirow{2}{*}{Method}} & \multicolumn{2}{c}{iNaturalist} & \multicolumn{2}{c}{SUN} & \multicolumn{2}{c}{Places} & \multicolumn{2}{c}{Textures} & \multicolumn{2}{c}{Average} \\

\cmidrule(r){2-3}  \cmidrule(r){4-5} \cmidrule(r){6-7} \cmidrule(r){8-9} \cmidrule(r){10-11}

\multicolumn{1}{c}{}  & AUROC$\uparrow$      & FPR$\downarrow$     & AUROC$\uparrow$    & FPR$\downarrow$     & AUROC$\uparrow$     & FPR$\downarrow$         & AUROC$\uparrow$   & FPR$\downarrow$    & AUROC$\uparrow$     & FPR$\downarrow$      \\
\hline
\multicolumn{11}{c}{\textbf{uni-modal OOD detection methods}}\\ 
\hline
\multicolumn{1}{c}{MSP~\cite{hendrycks2016baseline}}  
&77.74 &74.57 &73.97 &76.95 &74.84 &73.66 &72.18 &79.72 &74.68 &76.22  \\
\multicolumn{1}{c}{MaxLogit~\cite{hendrycks2022scaling}}  
&88.03 &60.88 &91.16 &44.83 &88.63 &48.72 &87.45 &55.54 &88.82 &52.49  \\
\multicolumn{1}{c}{Energy~\cite{liu2021energybased}}  
&87.18 &64.98 &91.17 &46.42 &88.22 &50.39 &87.33 &57.40 &88.48 &54.80  \\
\multicolumn{1}{c}{ReAct~\cite{sun2021react}}  
&86.87 &65.57 &91.04 &46.17 &88.13 &49.88 &87.42 &56.85 &88.37 &54.62  \\
\multicolumn{1}{c}{ODIN~\cite{liang2020enhancing}}  
&57.73 &98.93 &78.42 &88.72 &71.49 &85.47 &76.88 &87.80 &71.13 &90.23 \\
\multicolumn{1}{c}{KNN~\cite{sun2022knn}}  
&94.52 &29.17 &\textbf{92.67} &35.62 &91.02 &39.61 &85.67 &64.35 &90.97 &42.19 \\
\hline
\multicolumn{11}{c}{\textbf{multi-modal OOD detection methods}}\\ 
\hline
\multicolumn{1}{c}{MCM~\cite{ming2022delving}}  
&94.40 &32.18 &92.27 &39.29 &89.82 &44.92 &85.99 &58.03 &90.62 &43.61 \\
\multicolumn{1}{c}{NPOS~\cite{tao2023nonparametric}}  
&96.19 &16.58 &90.44 &43.77 &89.44 &45.27 &\textbf{88.80} &\textbf{46.12} &91.22 &37.93 \\
\multicolumn{1}{c}{Dai et al.~\cite{dai2023exploring}}  
&95.54 &22.88 &92.60 &\textbf{34.29} &89.87 &41.63  &87.71 & 52.02 &91.43 &37.71 \\
\bottomrule
\multicolumn{1}{c}{SimLabel-H}  
& 94.24 & 30.06& 89.99 & 51.07& 86.15 & 58.62& 81.03 & 72.09& 87.86& 52.96 \\
\multicolumn{1}{c}{SimLabel-L}  
& 96.15 & 19.13& 88.40 & 50.13& 91.42 & 45.01& 86.57 & 56.70& 90.64& 42.74\\
\multicolumn{1}{c}{SimLabel-I}  
& \textbf{96.74} & \textbf{15.28}& 90.35 & 42.84& \textbf{93.45} & \textbf{34.07}& 87.07 & 53.65& \textbf{91.90}& \textbf{36.46} \\

\bottomrule

\end{tabular}
\end{minipage} }
\end{table*}

\begin{table*}[t]
\footnotesize
\centering
\setlength{\tabcolsep}{2.75mm}
\renewcommand{\arraystretch}{1.1}
\caption{OOD detection results on various fine-grained datasets comparing with MCM where ID dataset is CUB-200\cite{WahCUB_200_2011}, Food-101\cite{bossard14food101}, Oxford-IIIT Pet\cite{oxford-pet} and Stanford Cars\cite{standford-cars}.
}
\label{table:fine-grained-OOD}
\begin{tabular}{lcccccccccccc}
\bottomrule 
\multicolumn{1}{c}{\multirow{2}{*}{ID Dataset}} & \multicolumn{1}{c}{\multirow{2}{*}{Method}} & \multicolumn{2}{c}{iNaturalist} & \multicolumn{2}{c}{SUN} & \multicolumn{2}{c}{Places} & \multicolumn{2}{c}{Textures} & \multicolumn{2}{c}{Average} \\

\cmidrule(r){3-4}  \cmidrule(r){5-6} \cmidrule(r){7-8} \cmidrule(r){9-10} \cmidrule(r){11-12}

\multicolumn{1}{c}{}  & \multicolumn{1}{c}{}  & AUROC$\uparrow$      & FPR$\downarrow$     & AUROC$\uparrow$    & FPR$\downarrow$     & AUROC$\uparrow$     & FPR$\downarrow$         & AUROC$\uparrow$   & FPR$\downarrow$    & AUROC$\uparrow$     & FPR$\downarrow$      \\
\hline 

\multicolumn{1}{c}{\multirow{2}{*}{CUB200}}  
& MCM         & 98.43 & 8.68& 99.07 & 4.94& 98.59 & 6.45& 99.05 & 4.70& 98.79& 6.19\\
& SimLabel-I    & \textbf{99.50} & \textbf{2.25}& \textbf{99.49} & \textbf{2.72}& \textbf{99.20} & \textbf{3.50}& \textbf{99.79} & \textbf{0.80}& \textbf{99.49}& \textbf{2.32}\\

\multicolumn{1}{c}{\multirow{2}{*}{Food101}}  
& MCM         & 99.39 & 1.81& 99.31 & 2.71& 99.07 & 4.01& 98.03 & 6.13& 98.95& 3.67\\
& SimLabel-I    & \textbf{99.54} & \textbf{0.98}& \textbf{99.42} & \textbf{2.11}& \textbf{99.28} & \textbf{2.90}& \textbf{98.22} & \textbf{5.30}& \textbf{99.12}& \textbf{2.82}\\

\multicolumn{1}{c}{\multirow{2}{*}{Pets}}  
& MCM         & 99.32 & 2.78& 99.75 & 0.93& 99.65 & 1.62& 99.78 & 1.01& 99.62& 1.59\\
& SimLabel-I    & \textbf{99.65} & \textbf{0.57}& \textbf{99.93} & \textbf{0.03}& \textbf{99.81} & \textbf{0.46}& \textbf{99.66} & \textbf{0.76}& \textbf{99.76}& \textbf{0.46}\\
\multicolumn{1}{c}{\multirow{2}{*}{Cars}}  
& MCM         & 99.79 & 0.09& \textbf{99.97} & \textbf{0.02}& \textbf{99.89} & \textbf{0.30}& \textbf{99.97} & \textbf{0.02}& 99.90& 0.11\\
& SimLabel-I    & \textbf{99.86} & \textbf{0.02}& 99.94 & 0.04& 99.87 & 0.33& 99.96 & \textbf{0.02}& \textbf{99.91}& \textbf{0.10}\\
\bottomrule 
\end{tabular}
\end{table*}

\textbf{Comparison with baselines.}
We conduct comprehensive OOD evaluation on the ImageNet-1k benchmark and compare our proposed method, namely SimLabel, with other existing OOD detection methods in Table~\ref{table:imagenetOOD}. 
The methods we compare can be divided into two categories: uni-modal OOD methods and multi-modal OOD methods. Specifically, the methods we compare include various multi-modal OOD methods based on MCM~\cite{ming2022delving, dai2023exploring} and several traditional uni-modal OOD detection methods including MSP~\cite{hendrycks2016baseline}, MaxLogit~\cite{hendrycks2022scaling}, Energy~\cite{liu2021energybased}, ReAct~\cite{sun2021react}, ODIN~\cite{liang2020enhancing} and KNN~\cite{sun2022knn}. Given the different similar classes construction strategies in Sec.~\ref{sec:choise_Sim_Class}, we show performance of
SimLabel-H, SimLabel-L and SimLabel-I, respectively in Table~\ref{table:imagenetOOD}.
Based on image-text alignment, we find
SimLabel-I performs the best across the proposed three different similar classes generation strategies, which makes sense given the extra multimodal information. Notably, although SimLabel-H and SimLabel-L perform overall inferior than SimLabel-I, they still achieve comparable performance with the SOTA methods. Notably, our proposed method, \textbf{SimLabel-I}, establishes a robust and discriminative image-class pairing which is MSP-based. Consequently, we focus our comparisons primarily on existing state-of-the-art MSP-based methods. 
On average, as a post-hoc method, our SimLabel-I, using the CLIP model with ViT-B-16 and similar classes generated by image-text alignment, demonstrates significant enhancements of 0.47\% and 1.25\% in terms of AUROC and FPR95 concerning formal \cite{dai2023exploring}. 

\textbf{OOD detection of SimLabel on fine-grained datasets.} Following the setup from MCM, we also explore the performance of SimLabel on small fine-grained datasets to evaluate our method's generalization ability.
Specifically, we utilize CUB-200~\cite{WahCUB_200_2011}, Food-101~\cite{bossard14food101}, Oxford-IIIT Pet~\cite{oxford-pet}, and Stanford Cars~\cite{standford-cars} as diverse ID datasets, reflecting a broad range of scenes and semantics. For OOD datasets, we employ iNaturalist~\cite{vanhorn2018inaturalist}, SUN~\cite{SUN}, Places~\cite{zhou2016places}, and Texture~\cite{cimpoi2013describing} following the setting in MCM~\cite{ming2022delving}. We show model performance in Table~\ref{table:fine-grained-OOD}, where our consistent better performance on various ID datasets indicates superior generalization ability our method comparing MCM~\cite{ming2022delving}.
Note that all the experimental results are evaluated using CLIP model based on ViT-B-16, where our proposed method SimLabel-I demonstrates significant improvement in OOD detection on most fine-grained ID datasets and OOD datasets.

\textbf{OOD detection of SimLabel on hard OOD detection tasks}.
The superior performance of SimLabel-I highlights its effectiveness in handling large-scale ID datasets, particularly when the similar classes exhibit strong semantic correlations. However, its applicability to smaller ID datasets remains uncertain. 
To address this, we investigate the performance of SimLabel-I on challenging OOD detection tasks, following the settings of MCM. As shown in Table~\ref{tab:hard-OOD}, our study focuses on semantic-hard OOD detection due to the lack of detailed guidelines from MCM for generating spurious OOD samples. To evaluate SimLabel-I, subsets of ImageNet (ImageNet-10 and ImageNet-20) were created, and these subsets were alternately used as ID and OOD datasets for testing.
The results presented in Table~\ref{tab:hard-OOD} demonstrate that SimLabel-I significantly outperforms MCM in hard OOD detection tasks. This finding underscores SimLabel-I's superior capability in distinguishing semantic-hard OOD samples. Furthermore, the effectiveness of SimLabel-I on smaller ID datasets showcases its remarkable generalization ability.

\begin{figure*}[!t]
\begin{minipage}[t]{\textwidth}
\begin{minipage}[t]{0.5\textwidth}
\makeatletter\def\@captype{table}
\scriptsize
\centering
\setlength{\tabcolsep}{3.2mm}
\renewcommand{\arraystretch}{1.2}
\caption{\textbf{Zero-shot OOD detection performance comparison on hard OOD detection tasks.} Following MCM~\cite{ming2022delving}, we use the subsets of ImageNet-1k\cite{imagenet2009} for testing the performance of SimLabel on hard OOD detection task. }
\label{tab:hard-OOD}
\begin{tabular}{cccccc}
\bottomrule 
ID dataset   & OOD dataset & Method   & AUROC$\uparrow$   & FPR95$\downarrow$ \\ \hline
\multicolumn{1}{c}{\multirow{2}{*}{ImageNet-10}} & \multicolumn{1}{c}{\multirow{2}{*}{ImageNet-20}} & MCM & 98.71 & 5.00 \\
\multicolumn{1}{c}{}  & \multicolumn{1}{c}{} & SimLabel-I & \textbf{99.30} & \textbf{3.20}\\
\multicolumn{1}{c}{\multirow{2}{*}{ImageNet-20}} & \multicolumn{1}{c}{\multirow{2}{*}{ImageNet-10}} & MCM & 97.88 & 17.40 \\
\multicolumn{1}{c}{}  & \multicolumn{1}{c}{} & SimLabel-I & \textbf{98.43} & \textbf{12.00}\\
\bottomrule 

\end{tabular}
\end{minipage}
\begin{minipage}[t]{0.5\textwidth}
\vspace{-0.1em}
\makeatletter\def\@captype{table}
\centering
\scriptsize
\setlength{\tabcolsep}{6mm}
\renewcommand{\arraystretch}{1.2}
\caption{\textbf{Zero-shot OOD detection with various VLM architectures other than CLIP.} We use the average performance of ImgeNet-100 (ID dataset) vs. four common OOD datasets: iNaturalist~\cite{vanhorn2018inaturalist}, SUN~\cite{SUN}, Places~\cite{zhou2016places}, and Texture~\cite{cimpoi2013describing}.} 
\label{table:model_ablation}
\begin{tabular}{cccccccc}
\bottomrule 
Architecture   & Method   & AUROC$\uparrow$   & FPR95$\downarrow$ \\ \hline
\multicolumn{1}{c}{\multirow{2}{*}{AltCLIP}} &  MCM & 83.40 & 71.70 \\
\multicolumn{1}{c}{}  & SimLabel-I & \textbf{84.66} & \textbf{65.13}\\
\multicolumn{1}{c}{\multirow{2}{*}{GroupViT}} & MCM & 69.45& 82.38 \\
\multicolumn{1}{c}{}  & SimLabel-I & \textbf{73.57} & \textbf{80.38}\\
\bottomrule 

\end{tabular}

\end{minipage}
\end{minipage}
\end{figure*}

\subsection{Discussion}\label{sec:dis}

\begin{figure*}
    \centering
    \includegraphics[width = 1.0\textwidth]{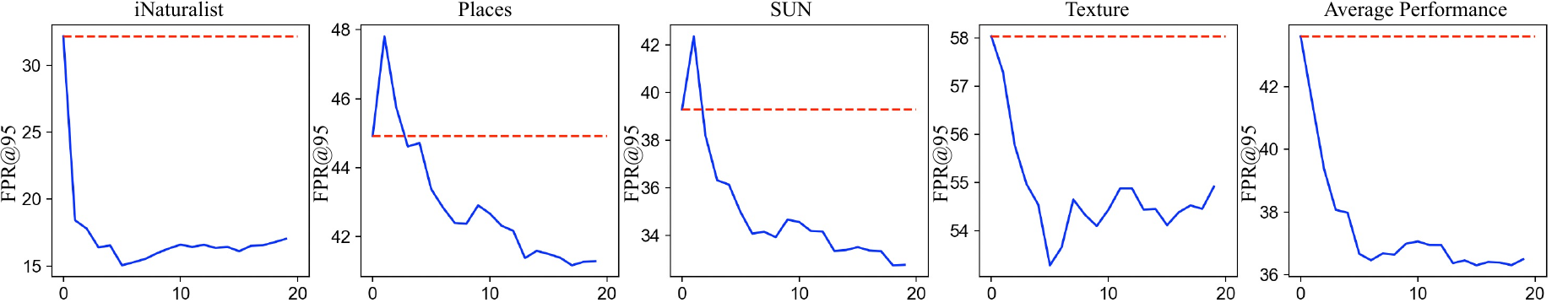}
    \caption{This figure illustrates how the FPR@95 changes with different choices on the number of similar classes ($k$) for each class in SimLabel-I Score on ImageNet-1k benchmark\cite{hendrycks2016baseline}. The x-axis and y-axis are the $k$ values and FPR@95 performance respectively. Additionally, the red dashed line is the MCM score result serving as a baseline for comparison.}
    \label{fig: Experiment with Similar classes}
\end{figure*}

\textbf{A small set of similar classes is sufficient.} 
We provide an empirical study to show that when generating similar classes, a small set of similar classes is sufficient. The similar classes generation mentioned in Sec.~\ref{sec:choise_Sim_Class} are based on selecting labels with top-$k$ similarities. If $k$ is too small, some representative classes may be overlooked, while a high $k$ value may results in the selection of non-semantically related labels (see Fig.~\ref{fig: Similar Classes Generated Visual}). 
Here, we empirically analyze the effectiveness of the number of similar classes in Fig.~\ref{fig: Experiment with Similar classes}. On average, a small set of similar classes within each class can significantly improve the effectiveness of OOD detection performance compared to the conventional MCM score. However, as the number of similar classes increases, the improvement in OOD detection tends to plateau. This explains our chosen of a relatively small
size of similar classes with $k=6$ in this paper.

\textbf{The weight of image \& similar-class-label similarity should be moderate.}
The class label $l_c$ plays a dominant role in representing class prototype and we provide an empirical study to show that the weight of image \& similar-class-label similarity should be moderate. We vary the hyper-parameter $\alpha$ within a wide range in Eq.~\ref{equa:affinity_simclass} and conduct the OOD detection with SimLabel-I score.
The experiment uses ImageNet-1k as ID and iNaturalist~\cite{vanhorn2018inaturalist}, SUN~\cite{SUN}, Places~\cite{zhou2016places}, and Texture~\cite{cimpoi2013describing} as various OOD datasets. The average performance over four OOD datasets is shown in Table~\ref{tab:alpha}.
The results indicate that either excessive emphasis on similar classes or their disregard is harmful to OOD detection.
Empirically, the optimal weight is found to be around $\alpha = 1$, indicating roughly equal contributions from image \& class-label similarity and image \& similar-classes-label affinity. 

\textbf{SimLabel on various VLM architectures} 
We conduct our main experiments based on CLIP-B/16~\cite{radford2021learning} while it is important to verify how SimLabel works on various VLM architectures. We provides additional experiments in investigating the effectiveness of SimLabel based on various VLM architectures including AltCLIP~\cite{chen2022altclip} and GroupViT~\cite{xu2022groupvit} which are two common-used VLM architectures. The experimental results are shown in Table~\ref{table:model_ablation} where we mainly compare with the MCM~\cite{ming2022delving} score. The results indicate that our method SimLabel-I significantly outperforms MCM based on various VLM architectures.

\section{A closer look at SimLabel}\label{Sec:analysis}

\subsection{SimLabel builds a robust and discriminative image-class pairing}

SimLabel enhances the separability between ID and OOD data by establishing a robust image-class pairing mechanism, leveraging consistency measurements across similar classes. As has benn outlined in Section~\ref{Sec:SimLabel}, for any ID image \( x \in \mathcal{X} \) and label set \( \mathcal{L} \), the OOD score is determined by the maximum matching score, and the class prediction is derived naturally using Eq.~\ref{equa:affinity_simclass} with the SimLabel score.
As maintaining high ID classification accuracy is a key factor in evaluating the effectiveness of our method, as outlined in Section~\ref{Sec:preliminaries}, a natural question arises: does this multi-label affinity hinder the model’s ability to learn discriminative image-class pairings?

To answer this question, we have conducted zero-shot visual classification using SimLabel.
To evaluate the impact of SimLabel on ID classification accuracy, we perform zero-shot visual classification on several datasets, including the large-scale ImageNet dataset, ImageNetV2~\cite{imagenet2009, recht2019imagenetv2}, and the small-scale fine-grained CUB-200 dataset~\cite{WahCUB_200_2011}. The results, presented in Table~\ref{table:ZSL-classification}, demonstrate that our method, which pairs images with similar class labels, improves ID classification accuracy compared to traditional zero-shot visual classification methods based on Eq.~\ref{equa:CLIP_classify}. These findings suggest that SimLabel enhances the robustness and discriminative power of image-class prototype matching, thereby not only improving ID/OOD separability but also preserving high ID classification accuracy.

\begin{figure*}[!t]
\begin{minipage}[t]{\textwidth}
\begin{minipage}[t]{0.5\textwidth}
\makeatletter\def\@captype{table}
\scriptsize
\centering
\setlength{\tabcolsep}{2.3mm}
\renewcommand{\arraystretch}{1.2}
\caption{\textbf{The influence of $\alpha$ on SimLabel-L and SimLabel-I.} We use the average performance of ImgeNet-1k (ID) vs. four common OOD datasets: iNaturalist~\cite{vanhorn2018inaturalist}, SUN~\cite{SUN}, Places~\cite{zhou2016places}, and Texture~\cite{cimpoi2013describing}. }
\label{tab:alpha}
\begin{tabular}{ccccccccc}
\bottomrule 
$\alpha$   &0 & 0.1   & 0.5   & 1   & 5   & 10   & 100            \\ \hline
SimLabel-I &43.61 & 41.63 & 37.73 & 36.46 & 40.27 & 42.05 & 50.92  \\
SimLabel-L &43.61 & 42.24 & 41.13 & 42.74 & 51.70& 54.58 & 62.83  \\
\bottomrule 

\end{tabular}

\end{minipage}
\begin{minipage}[t]{0.5\textwidth}
\makeatletter\def\@captype{table}
\centering
\scriptsize
\setlength{\tabcolsep}{5.5mm}
\renewcommand{\arraystretch}{1.2}
\caption{\textbf{Zero-shot visual classification performance on various datasets.} We demonstrate the accuracy of SimLabel-I and CLIP-B/16 in doing prediction over several common datasets cub200~\cite{WahCUB_200_2011}, ImageNet~\cite{imagenet2009} and ImageNetV2~\cite{recht2019imagenetv2}} 
\label{table:ZSL-classification}
\begin{tabular}{cccccccc}
\bottomrule 
Datasets     & ImageNet         & ImageNetV2         & cub200                \\ \hline
CLIP-B/16        & 66.60            & 60.61              & 55.71                  \\
SimLabel-I & 67.88            & 61.29              & 57.56                  \\ 

\bottomrule 

\end{tabular}

\end{minipage}
\end{minipage}
\end{figure*}

\subsection{The effectiveness of isolated image \& similar-classes-label affinity.}
Our proposed method, the SimLabel score, operates on the assumption that, when measuring similarity between ID samples and the set of ID labels, other ID labels exhibit high similarity in addition to the ground truth class label $l_c$. In Sec.~\ref{sec:motivation}, we have provided a straightforward experiment, visualizing the similarity between a random image and various ID classes, highlighting the presence of similar classes. In this section, we extend the investigation from the perspective of OOD detection by conducting additional experiments to verify the inherent connections between different ID classes. 
In an ideal scenario, the affinity between an image and its similar-class labels should remain representative of the image-class pairing, even in the absence of the ground truth label $l_c$. To verify this, we reformulate the image-class prototype similarity defined in Eq.~\ref{Equa:simlael_equation} to consider only the similarity between the image and its similar-class labels. This reformulated affinity is computed as:
\[
\mathcal{A}(x, l_c) = \frac{\sum_{d \in \mathcal{D}(l_c)} \mathcal{M}(x, d)}{|\mathcal{D}(l_c)|}.
\]

If this assumption holds,
the similarity between the ID image and other ID labels would be indistinguishable, undermining the meaningful design of image-class affinity.
To evaluate this approach, we have applied the affinity metric for OOD detection using the similar classes generated for SimLabel-I on the ImageNet benchmark. The results, presented in Table~\ref{table:comparing_no_IDlabel} (denoted as SimLabel-S), show that while the performance of SimLabel-S is not as strong as SimLabel-I, its OOD detection capability still demonstrates the effectiveness of representing a class prototype without requiring the ground truth label. 
This finding verifies the assumption illustrated in Fig.~\ref{fig:motivation_sim_class} and provides robust support for the conceptual design of SimLabel.
\begin{table*}[t]
\centering

\setlength{\tabcolsep}{3.6mm}
\renewcommand{\arraystretch}{1.2}
\caption{Zero-shot OOD detection of SimLabel-S and SimLabel-I on ImageNet-1k benchmark following MCM~\cite{ming2022delving}. }
\label{table:comparing_no_IDlabel}
\begin{tabular}{lcccccccccc}
\bottomrule

\multicolumn{1}{c}{\multirow{2}{*}{Method}} & \multicolumn{2}{c}{iNaturalist} & \multicolumn{2}{c}{SUN} & \multicolumn{2}{c}{Places} & \multicolumn{2}{c}{Textures} & \multicolumn{2}{c}{Average} \\

\cmidrule(r){2-3}  \cmidrule(r){4-5} \cmidrule(r){6-7} \cmidrule(r){8-9} \cmidrule(r){10-11}

\multicolumn{1}{c}{}  & AUROC$\uparrow$      & FPR$\downarrow$     & AUROC$\uparrow$    & FPR$\downarrow$     & AUROC$\uparrow$     & FPR$\downarrow$         & AUROC$\uparrow$   & FPR$\downarrow$    & AUROC$\uparrow$     & FPR$\downarrow$      \\
\hline

\multicolumn{1}{c}{SimLabel-I}  
& 96.74 & 15.28& 90.35 & 42.84& 93.45 & 34.07& 87.07 & 53.65& 91.90& 36.46  \\
\multicolumn{1}{c}{SimLabel-S}  
& 95.13 & 25.50& 86.88 & 56.29& 90.59 & 50.84& 82.51 & 60.30& 88.78& 48.23 \\

\bottomrule
\end{tabular}
\end{table*}
\subsection{Interpretable visualization of SimLabel}
\begin{figure}
    \centering
    \includegraphics[width = 0.4\textwidth]{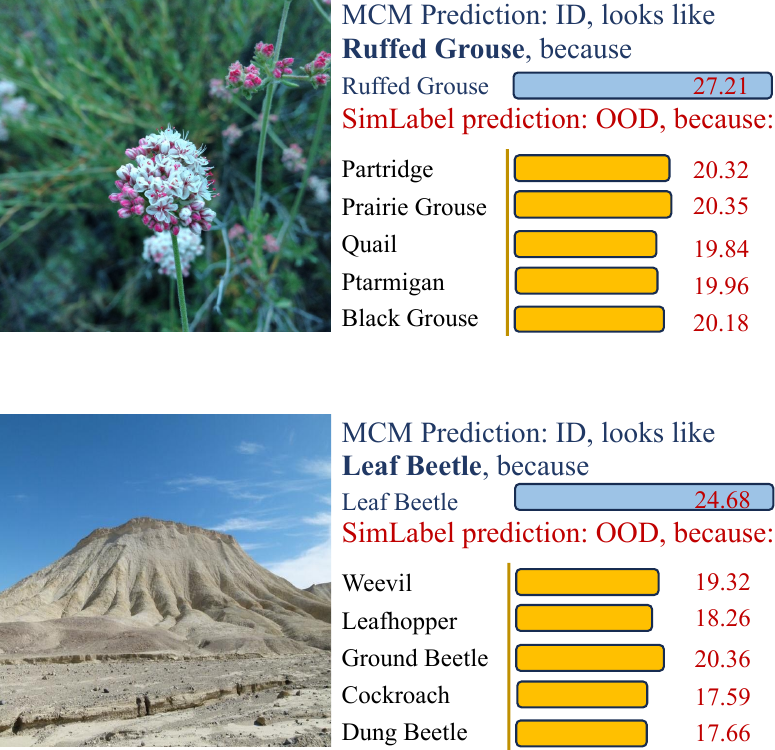}
    \caption{This figure presents two practical examples illustrating how our proposed SimLabel score improves OOD detection compared to the MCM baseline. In these examples, SimLabel successfully identifies the OOD images by leveraging their poor consistency with the similar classes.}
    \label{fig: intrepretable-visualizaiont}
\end{figure}

Building an image-class prototype using the alignment of images with similar-class labels offers an interpretable approach to enhancing image-class connections. To illustrate this interpretability, we provide a visualization in Fig.~\ref{fig: intrepretable-visualizaiont}. The figure highlights two OOD images that are incorrectly classified as ID by the MCM score from \cite{ming2022delving}. Our method, on the other hand, successfully detects these OOD images by leveraging consistency measurements with similar classes, demonstrating the advantage of SimLabel in capturing meaningful semantic relationships.

\subsection{Limitation analysis}\label{Sec:limitation}
The proposed method utilizes a \enquote{similar class pool} to compute a robust OOD score. A critical prerequisite for its effectiveness is that the distribution of ID classes is relatively uniform or balanced. 
In long-tailed scenarios, identifying informative similar classes for tail classes becomes significantly challenging compared to head classes. This imbalance presents substantial difficulties for the consistency-guided OOD detection approach. Specifically, the non-uniform distribution of class-wise similar classes hinders the accurate measurement of image-class affinity through consistency, which explains the observed failure cases of SimLabel-H in zero-shot OOD detection. Table~\ref{tab:hierarchical-set} illustrates subsets of hierarchical connections, highlighting the imbalance across super-classes and its limiting impact on OOD score construction.
One potential solution to this issue involves extending tail classes by incorporating additional sibling or child classes.

Another limitation of the proposed method lies in the design of image and similar-class-label similarity in Eq.~\ref{equa:affinity_simclass}, where an equal contribution is assumed for each similar class \( d \in \mathcal{D}(l_c) \). In real-world scenarios, semantic similarity between classes is not uniform, and assigning equal weights to all similar classes limits the effective utilization of semantic information. 
To address this, an extension of Eq.~\ref{equa:affinity_simclass} could be explored, incorporating a mechanism to accurately measure and weight class-wise similarity within the similar class pool. This enhancement would improve the method's ability to leverage semantic relationships and further refine the calculation of image-class affinity.

\section{Conclusion}
This paper presents SimLabel, a simple yet effective post-hoc method for multi-modal zero-shot OOD detection. SimLabel generates a set of class-wise similar labels that exhibit semantic relationships to each ID class. Through a series of experiments, we verify the existence of similar classes, demonstrating the inherent connections among various ID labels.
The fundamental assumption behind SimLabel is that an ID sample should consistently exhibit high similarity scores across its associated similar ID classes. This assumption forms the foundation of the proposed consistency-guided OOD detection framework. Specifically, SimLabel determines whether an image is ID or OOD by measuring and comparing its affinity towards both ID labels and their corresponding similar classes.
We propose three distinct strategies to generate similar classes: 1) selection based on text hierarchy, 2) selection using LLMs, and
3) utilizing pseudo-image-text pairing to align labels with visual information. We present extensive experiments and a comprehensive analysis of the three similar label generation strategies. These experiments are conducted across various OOD detection benchmarks, demonstrating the effectiveness of SimLabel-I, where similar classes are determined with image-text alignment. The extensive results highlight SimLabel-I's superiority in addressing challenging OOD detection tasks, showcasing its ability to outperform existing methods in complex scenarios.

{\small
\bibliographystyle{ieeetr}
\bibliography{Main_Paper}
}

\end{document}